# Anomaly Detection for Network Connection Logs

Swapneel Mehta

Dept. Of Computer Engineering,
D. J. Sanghvi College of Engineering
Mumbai, India
swapneel.mehta@djsce.edu.in

Prasanth Kothuri, Daniel Lanza Garcia

IT-DB Group
European Organisation for Nuclear Research
Geneva, Switzerland
{prasanth.kothuri, daniel.lanza}@cern.ch

**We leverage a streaming architecture based on ELK, Spark and Hadoop in order to collect, store, and analyse database connection logs in near real-time. The proposed system investigates outliers using unsupervised learning; widely adopted clustering and classification algorithms for log data, highlighting the subtle variances in each model by visualisation of outliers. Arriving at a novel solution to evaluate untagged, unfiltered connection logs, we propose an approach that can be extrapolated to a generalised system of analysing connection logs across a large infrastructure comprising thousands of individual nodes and generating hundreds of lines in logs per second.**

*Network Connection Logs, Anomaly Detection, Unsupervised Learning, Big Data Architecture, Clustering, Data Streaming*

## I.  INTRODUCTION

Anomaly detection has provided a classic problem statement across multifarious use-cases ranging from scientific observations to financial transactions. We define an anomaly as a single observation or a set thereof, that fails to conform to a group of properties exhibited by larger collections of such observations.

While anomalies are often tagged as undesirable in certain domains, they are representative of a highly specialised subset that provide insight into interesting phenomena within a system. Particularly in the domain of computer networks, intrusion detection and security systems, outliers can signify unusual activity critical for the health of a system. They form the most important part of monitoring activity, as spikes and dips can result in implications including attackers gaining access to the internal network, malware-initiated network scans, or hosts losing connectivity and crashing.

## II.  THE CERN NETWORK

The network of the European Organisation for Nuclear Research (CERN) comprises some 10,000 individual users and associated devices signing in both on-site as well as remotely into the system. The activity logs generated are monitored, analysed, and stored in order for meaningful insights to be generated and a historical archive of records to be maintained for future reference [1].

For an organisation working with experiments with the capacity to generate upto 30 petabytes of data each year**,** it is imperative to maintain the health of a network that can sustain such bandwidth on this scale with a high fault tolerance and extremely low probability for failure. The Worldwide Large Hadron Collider Computing Grid (WLCG) was set up around 2002 in order to distribute the processing load over a multi-tiered architecture across a global network of 42 countries. This includes a datacenter at the complex in Meyrin and the Wigner Research Centre in Budapest at the centre of all computation and data storage operations [2].

## III.  DATABASE SERVICES AT CERN

The Database Services Group at CERN is responsible for the administration and management of data from the experiments. It manages the assortment of critical services and web applications offered at CERN scale. This group is responsible for provision of an enterprise analytics infrastructure comprising Spark, Hadoop, Kafka and so on **[3] []**. The setup comprises of nearly 1,000 Oracle Databases, most of them being Real Application Clusters. With nearly 950TB of data files for production databases excluding replicas, and a logging system of 492TB growing at nearly 180TB annually, there is a need for a robust streaming architecture.

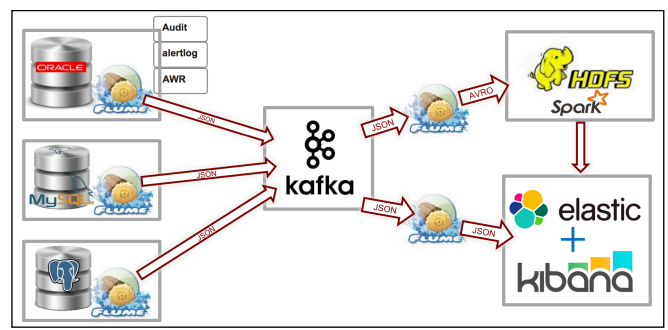

1.          Overview of the Data Pipeline for Streaming [5]

Such an architecture has been set up to allow for data streaming and storage. The aggregated log data from incoming database connection requests is streamed as a "notification" by Apache Flume Connectors to a Kafka buffer. This provides a highly flexible, configurable option and a containable memory footprint. It is ultimately stored in one of two ways:

- Temporary short-term storage on Elasticsearch and visualisation using Kibana to determine short-term anomalies in the database connections.

- Long-term storage on Hadoop Distributed File System (HDFS) in Parquet format (to meet compression requirements) that can be retrieved for analysis.

The architecture provides for a robust model that can permit near real-time streaming and visualisations. The monitoring encompasses notifications that include alerts, audits and performance metrics. A listener is attached to each database instance for the purpose of tracking connection requests as they come in. These are streamed via the buffer for storage as required for a short-term or long-term duration. Such a strategy has been proposed in [3].

## IV. Data Lake

The objective of this data lake is to build a central repository for database audit, performance metrics and logs with the goal of real-time analytics as well as offline analytics.

Further, a store such as this one presents an opportunity to investigate strategies built around anomaly detection, carry out capacity planning as well as troubleshooting.

### A. Connection Parameters

```
{
    "oracle_sid": "XXXX_YYYY",
    "listener_name": "listener_scan1",
    "database_type": "oracle",
    "producer": "oracle",
    "source_type": "listener",
    "hostname": "XXXXX.cern.ch",
    "flume_agent_version": "0.1.6-7.el6",
    "type": "dbconnection",
    "event_timestamp": "2017-09-27T04:45:27+0200",
    "CONNECT_DATA_SERVER": "DEDICATED",
    "CONNECT_DATA_SERVICE_NAME": "XXXr.cern.ch",
    "client_program": "python",
    "client_host": "pxxxxj2.cern.ch",
    "client_user": "merge",
    "client_protocol": "tcp",
    "client_ip": "137.100.100.167",
    "client_port": 38432,
    "type": "establish",
    "service_name": "XXXX.cern.ch",
    "return_code": 0
}
```

2.  Overview of the Data Pipeline for Streaming [5]

The log data serves as audit data, performance metrics and alerts, and comprises of fields utilised to build the feature vector. These logs are utilised to extract a useful subset of information from the system, and form the first stage in the preprocessing pipeline for building models for outlier detection among the connections.

### B. Data Ingestion

There are challenges faced with regard to scalability, when the data ingestion pipelines are set up:

1. The heterogenous nature of data sources that include databases, REST APIs, web sources and logs.

2. HDFS serves as a file store, not a database, thus some of the core features offered by a database system are not directly available and must be integrated using indirect means.

3. While HDFS offers a broad range of functionality there are certain limitations that do tend to impact the latency of the system.

There are a number of requirements for real-time data streaming proposed in [4] that we must be mindful of for the sake of scalability and low-latency. Some data sources and softwares are common across an array of network log streaming systems while they vary in other aspects including use-cases, latency, storage mechanism and scalability.

### C. Evaluation of Data Pipeline

There were a number of tests performed in order to evaluate the performance of the system. The major points of interest were data storage mechanism as well as the distributed messaging system within the proposed architecture:

1. Figures 3 and 4 show the results of the data storage format comparison for the Parquet and Avro formats. We pick Parquet because of the scan performance and low latency for analytical queries.

2.

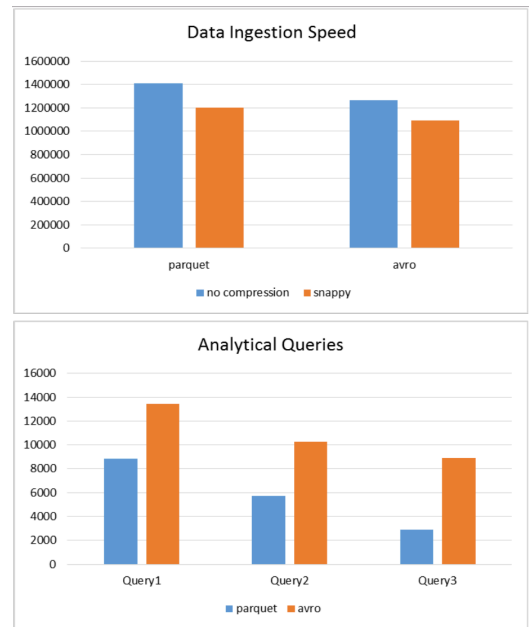

3.  Avro vs. Parquet performance over Analytical Queries

3. Figure 5 shows the results that were obtained when we benchmarked our systems with Flume-driven messaging versus a Kafka-driven messaging queue.

Our architecture is modelled using the results of these tests that provide a clearer idea of the scalability and extension of such a system over time. It has its own set of drawbacks but we minimise these by utilising best practices at scale.

The current CERN IT Monitoring and Logging architecture also faces a subset of issues pertaining to the increased luminosity of experiments implying the generation of greater volumes of data with each run of the Large Hadron Collider. However, there have been coordinated efforts targeted towards data acquisition and filtering, thus reducing both the computational load and storage requirements for the CERN Data Management Systems.

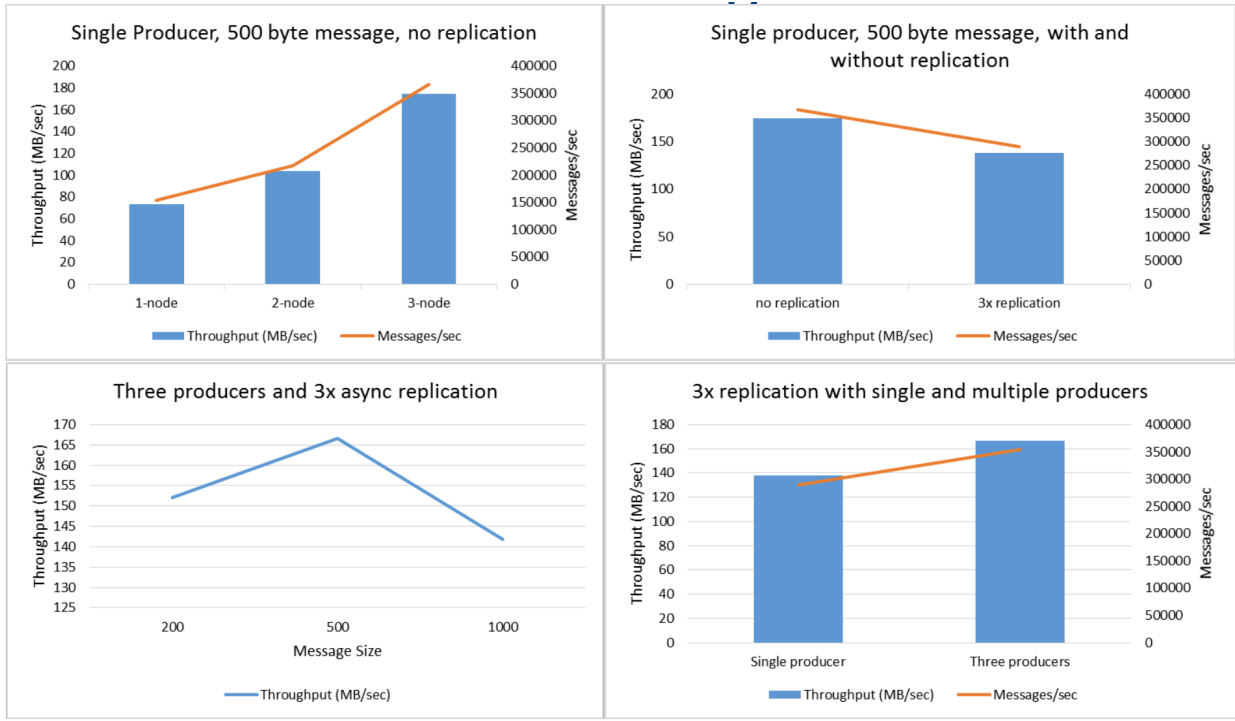

4.      Flume vs. Kafka Benchmarking on CERN Openstack Infrastructure - 4 virtual CPUs, 7.3 gigabytes RAM, and 100 gigabytes storage

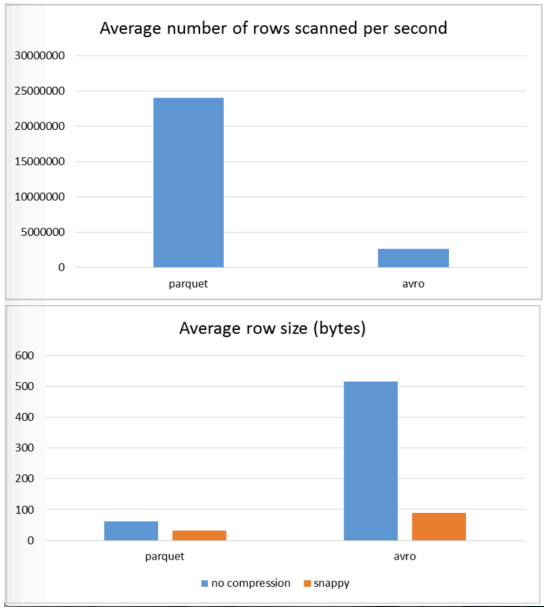

5.      Avro vs. Parquet performance over Scanning

## V.      ANOMALY DETECTION

The term 'outliers' refers to the subset of a series of data points that fail to conform to the conventional pattern that the rest of these points appear to satisfy.

Outliers have often been discarded as undesirable occurrences due to their nature of interfering with the insights and patterns obtained from known observations, however this very nature has now deemed them invaluable especially in the nature of fields where data acquisition and filtering is required to minimise the useful data points. The techniques proposed in [7] for dealing with outliers in high dimensional data present a simplified picture of why feature selection becomes increasingly important as the outliers become susceptible to noise in high dimensional data. We implement principal component analysis (PCA) as in [8] and singular vector decomposition (SVD) on cross-validated slices of our dataset in order to allow for dimensionality reduction.

Anomaly detection is a heavily studied subject [11. 12] involving not only supervised [14, 15] and unsupervised approaches [5, 13] but also rule-based systems [17]. Recent research into intrusion detection demonstrates that among the many techniques adopted include unsupervised approaches that forego the need for labeled training data. These solutions entail clustering of data based on different metrics including distance, density and so on, and will be the focus of our research into anomaly detection.

| Parameter | Model | |
| --- | --- | --- |
| | *Method* | *Contamination* |
| Distance | K-Nearest Neighbours | 2% |
| Density | Isolation Forests | 3% |
| Density | Local Outlier Factor | 5% |
| Classification | One-class SVM | 2% |

6.      Comparison of approaches adopted within the ensemble

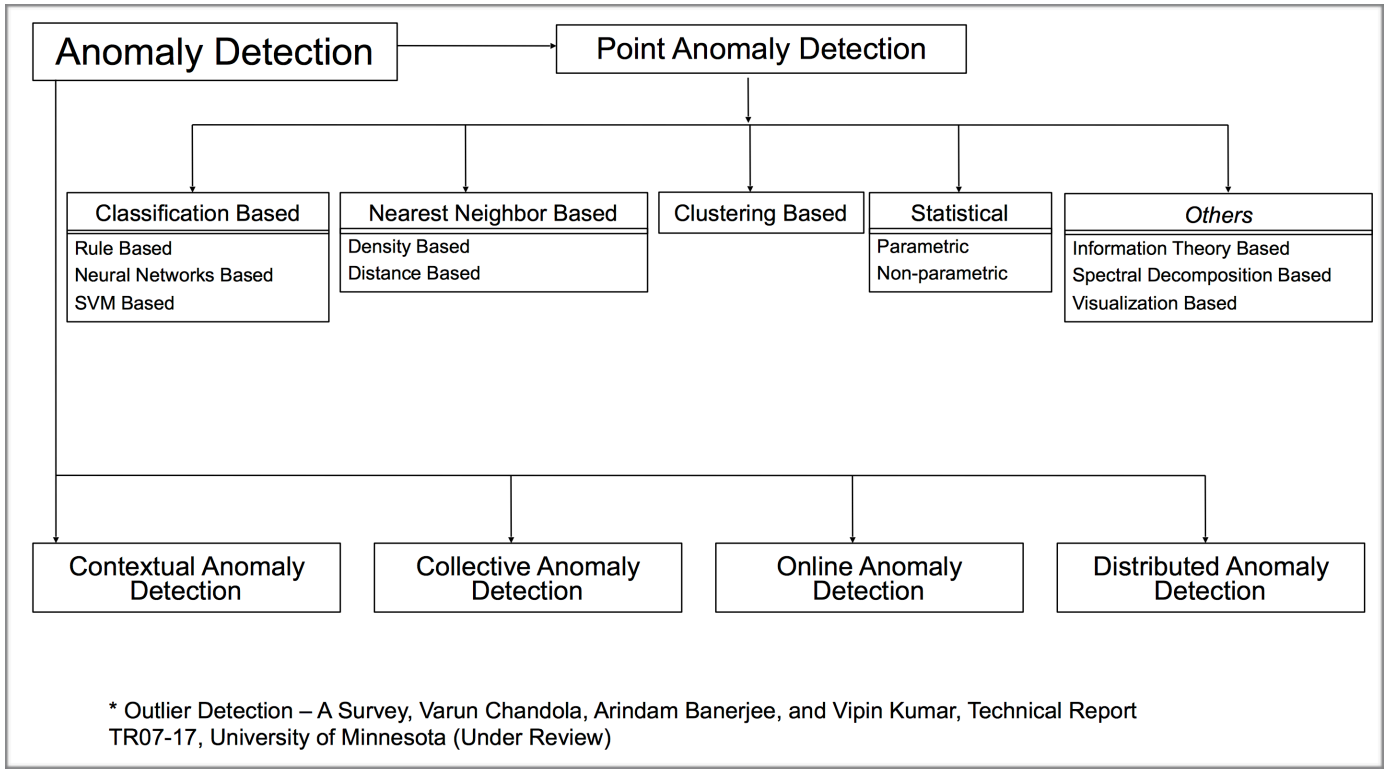

* Outlier Detection – A Survey, Varun Chandola, Arindam Banerjee, and Vipin Kumar, Technical Report TR07-17, University of Minnesota (Under Review)

7.      Survey of Unsupervised Anomaly Detection Approaches []

The table presents an overview of the test conditions that we utilised in order to evaluate the ensemble of algorithms for detecting anomalous database connections within the logs streamed into the data lake. The contamination refers to the subset of original data that we assumed to be contaminated and filtered out as outliers as part of training the models.

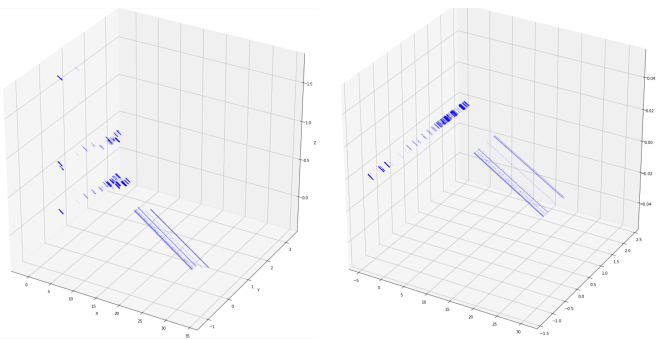

8.      Implementing SVD and PCA [7], reducing dimensionality to 3-dimensional (left) and 2-dimensional tuples for plotting.

## VI.    MODELS

The algorithms follow the paradigm of unsupervised learning due to the absence of any labeled training data and are distributed by the metric that they evaluate i.e. distance, density and classification-based methods as shown in the chart above.

However, the initial stages for this pipeline involved the preprocessing of data utilising PCA and SVD. The high-dimensional data was thus reduced to a smaller number of dimensions that could be then utilised for generating visualisations.

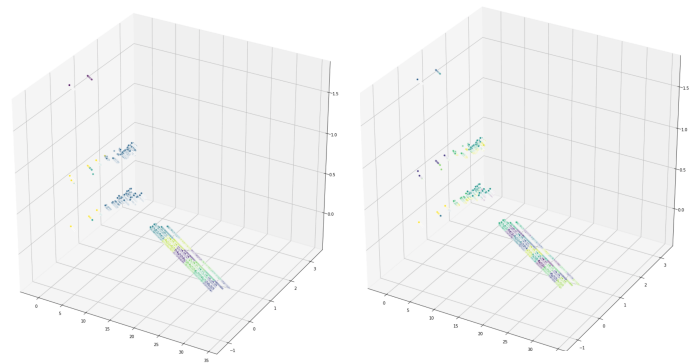

9.      K-means clustering allows us to visualise the data and arrive at a reasonable estimate for modalities.

We utilise an implementation of the k-means clustering algorithm proposed in [10] which presents a simplified picture of the number of modalities of the data, helpful when deciding metrics such as contamination, and in appreciating the evaluation presented by each individual model. In addition to evaluation metrics for unsupervised clustering methods such a silhouette.

### A.   K-Nearest Neighbours (kNNs)

Many strategies [5, 6, 9] for outlier detection have emphasised on clustering as a central element for their

approach(es). One of the elemental steps in clustering with a view to classifying anomalies, kNN clustering bases the grouping on a distance-oriented metric. Specifically, it calculates the Euclidean distance of a given point from the set of points constituting the "neighbours" of said point, and an inherent voting system ensures the point falls into the most likely category of similar connection logs, in this case.

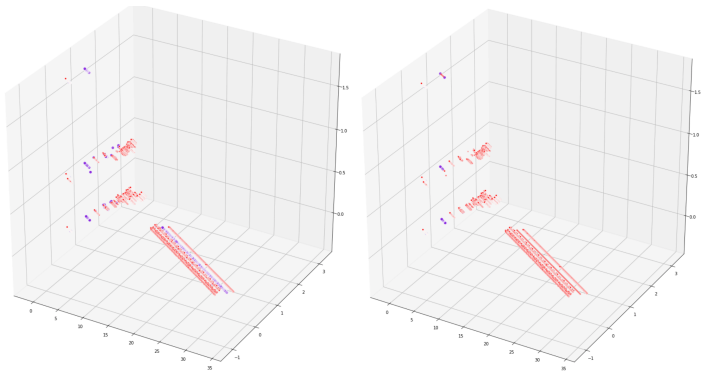

10. K-nearest neighbours with number of anomalies (plotted in violet) filtered and optimised from left to right.

We optimise the results by implementing a recurring filter that uses a predetermined threshold for eliminating false-positives. Simply put, we recursively iterate the anomalies to filter them to 2% of the original dataset, thus excluding the vast majority of the normal data wrongly tagged as anomalous.

### B. Isolation Forests

A near linear time-complexity algorithm, Isolation Forests or iForests similar to the ones proposed in [18] allow the adoption of an isolation based approach that can achieve performance similar to random forests and local outlier factor based methods to detect anomalies.

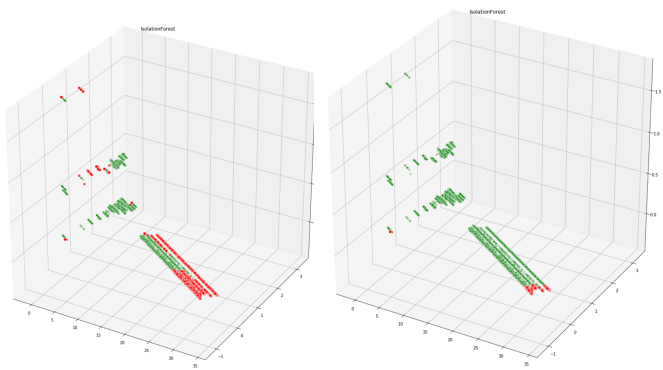

11. Isolation forests with outlier detection (outliers in red) optimised from left to right.

In figure 11, the optimisation carried out reduces the number of anomalies by diminishing the explicit threshold for contamination as set within the data. This method is typically useful for datasets with high-dimensionality with large numbers of irrelevant attributes due to its low computational complexity compared to other algorithms.

### C. Local Outlier Factor

Outlier detection has often been considered a binary classification problem. However, in [19], the authors propose a local outlier factor that provides a degree of isolation with regards to its neighbourhood. This is interesting as it lays emphasis on the local environment of a data point and not the global density distribution.

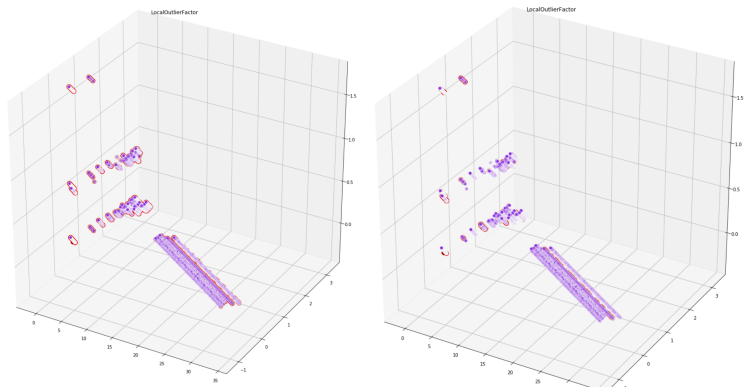

12. Local outlier factor showing detected outliers (outliers encircled in red) that are optimised using a threshold from left to right.

As in the previous cases, we assume the training data to be contaminated in itself since it is illogical to assume the captured logs comprise of entirely legal activity within the system. The reduction in red circled data points denotes a marked reduction in false-positives for the algorithm, to put simply.

### D. One-class Support Vector Machine

Generally constituting the bulk of methods adopted to address supervised multiclass classification problems, one-class support vector machines (OCSVMs) are a special class of Support Vector Machines that allow the analysis of an input array with no class labels and generate a soft boundary for its analysis [16].

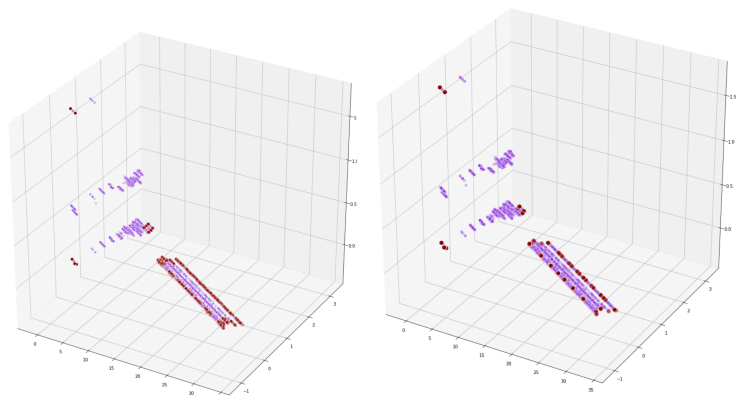

13. OCSVM with novelty detection (outliers in maroon) optimised from left to right.

We extrapolate the application of OCSVMs to novelty detection that uses its kernel function to map the data into a feature space and thereby analyse the outliers. By controlling the contamination metric in the original data, we arrive at a reduced false-positive detection in the system.

## VII. EVALUATION

To arrive at a performance evaluation for such a system we must consider firstly that there is no absolute figure of merit and due to the lack of tagged data, there is no possible confusion matrix to be derived. However, there are practices that can calculate the efficiency achieved by measuring the similarity of a data point to the rest of the cluster. One such approach utilises the silhouette metric often used in unsupervised learning.

While it has been proposed that randomised search can often yield results comparable to grid search in cases where time and space complexity is an issue [20], we are not constrained by either and hence can implement such an approach in an attempt to optimise the hyperparameters: here specifically the contamination metric across models.

We find that the maximum silhouette score of 0.35 across ensembles was obtained at the contamination values shown in figure 6. This serves as one part of the evaluation for our experiment. The other, more manual portion involved the cross-verification of the detected anomalies against security incidents within the system. We find that there are three categories that these detected results fall into:

1. Malware triggering reconnection requests from hosts thus resulting in system load and increased transfer of data between the server and specific host.

2. User login outside the regular pattern e. g. working on a holiday; since it fits the definition of an anomaly, such an event is duly detected and flagged within the system.

3. User requesting multiple resources not necessarily useful to successful operation and without a history of access for such resources.

While the experiment serves as a proof-of-concept for the introduction of a formal system, these results are encouraging especially in the absence of any existing system within the CERN network to detect or flag anomalous connections. More importantly, such an apparatus can be generalised to other applications involving the analysis of large chunks of log data or data in similar formats, using the presented architecture, in order to increase the efficiency and improve the performance of the system.